\documentclass[sigconf]{acmart}
\AtBeginDocument{%
  \providecommand\BibTeX{{%
    \normalfont B\kern-0.5em{\scshape i\kern-0.25em b}\kern-0.8em\TeX}}}

\acmYear{2021}\copyrightyear{2021}
\setcopyright{acmcopyright}
\acmConference[ACM CHIL '21]{ACM Conference on Health, Inference, and Learning}{April 8--10, 2021}{Virtual Event, USA}
\acmBooktitle{ACM Conference on Health, Inference, and Learning (ACM CHIL '21), April 8--10, 2021, Virtual Event, USA}
\acmPrice{}
\acmDOI{10.1145/3450439.3451869}
\acmISBN{978-1-4503-8359-2/21/04}

\usepackage{booktabs} 
\usepackage{dsfont}
\usepackage{tikz}
\usetikzlibrary{bayesnet}
\usepackage{enumitem}
\usepackage{amsmath}
\usepackage{amsfonts} 
\usepackage{bbm}
\usepackage{bm}
\usepackage{setspace}
\usepackage{subcaption}

\begin{document}

\title[Learning to Predict with Supporting Evidence]{Learning to Predict with Supporting Evidence: \\ Applications to Clinical Risk Prediction}

\author{Aniruddh Raghu}
\email{araghu@mit.edu}
\affiliation{%
  \institution{Massachusetts Institute of Technology}
}

\author{John Guttag}
\affiliation{%
  \institution{Massachusetts Institute of Technology}
}

\author{Katherine Young}
\affiliation{%
  \institution{Massachusetts Institute of Technology \\ Harvard Medical School}
}

\author{Eugene Pomerantsev}
\affiliation{%
  \institution{Massachusetts General Hospital}
}

\author{Adrian V. Dalca}
\affiliation{%
  \institution{Massachusetts Institute of Technology \\ Massachusetts General Hospital Harvard Medical School}
}

\author{Collin M. Stultz}
\affiliation{%
  \institution{Massachusetts Institute of Technology \\ Massachusetts General Hospital Harvard Medical School}
}

\renewcommand{\shortauthors}{A. Raghu, et al.}

\begin{abstract}
The impact of machine learning models on healthcare will depend on the degree of trust that healthcare professionals place in the predictions made by these models. In this paper, we present a method to provide individuals with clinical expertise with domain-relevant evidence about why a prediction should be trusted. We first design a probabilistic model that relates meaningful latent concepts to prediction targets and observed data. Inference of latent variables in this model corresponds to both making a prediction \textit{and} providing supporting evidence for that prediction. We present a two-step process to efficiently approximate inference: (i) estimating model parameters using variational learning, and 
(ii) approximating \textit{maximum a posteriori} estimation of latent variables in the model using a neural network, trained with an objective derived from the probabilistic model.
We demonstrate the method on the task of predicting mortality risk for patients with cardiovascular disease. Specifically, using electrocardiogram and tabular data as input, we show that our approach provides appropriate domain-relevant supporting evidence for accurate predictions. 
\end{abstract}



\maketitle
\section{Introduction}

There is significant interest in using machine learning (ML) models in high-stakes domains such as medicine, where human experts use model predictions to inform decisions~\cite{cai2019human,dearteaga2020human,topol2019high}.  When used by a clinician, the utility of an ML model depends not only on the accuracy of its predictions, but also on how much trust the clinician places in the prediction \citep{stultz2019advent}.

Building trust in predictions can be tackled in different ways, including (1) providing statistical arguments to reflect the level of certainty in a prediction~\citep{rasmussen2003gaussian,myers2020identifying} (2) providing an explanation of how a model reached its prediction, either through understanding important input features that led to the model output, or by using an inherently explainable model~\cite{melis2018towards,sundararajan2017axiomatic}; and (3)~offering \textit{domain-specific supporting evidence} about the prediction~\cite{hind2019ted}.

\begin{figure}[t!]
\centering
  \includegraphics[width=\linewidth]{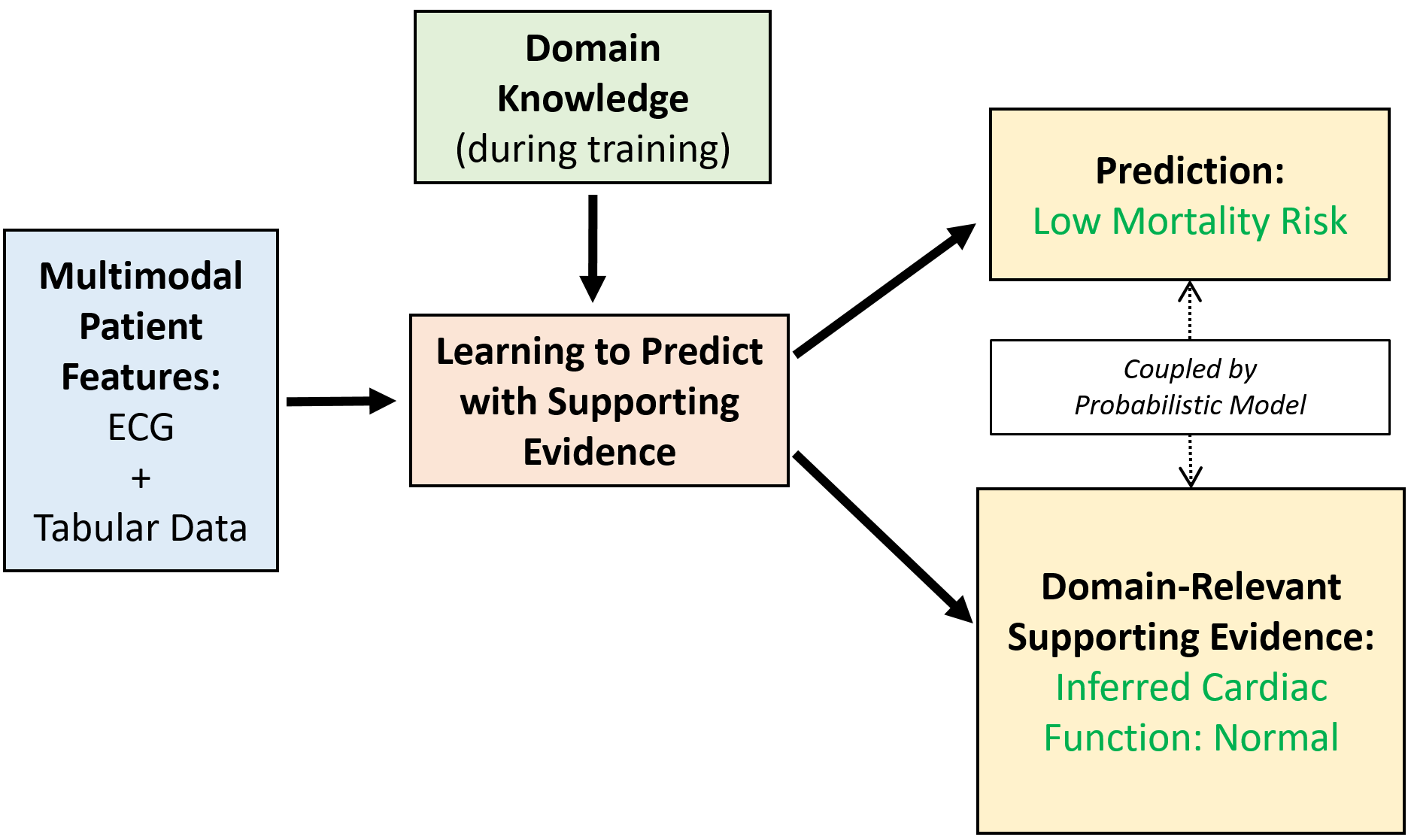}
  \caption{\textit{Learning to Predict with Supporting Evidence.} We propose a model that produces both a prediction and accompanying \textit{domain-relevant supporting evidence} for that prediction: inferred concepts that are clinically meaningful and can assist a clinician in deciding how to act on the prediction. Consistency of the prediction and supporting evidence is ensured using a probabilistic model and domain knowledge.}
  \label{fig:model-overview}
\end{figure}

For models used by domain experts with a limited understanding of machine learning, prior work has emphasised the importance of providing \textit{domain-relevant supporting evidence} for predictions~\citep{kulesza2013too,miller2019explanation}.
In medicine, presenting relevant clinical information that supports a model prediction can be crucial, since clinicians draw significantly on medical knowledge when making decisions~\citep{stultz2019advent,tonekaboni2019clinicians}. For example, a cardiologist might deem a patient to be at high risk of death (prediction) because of their low cardiac output (domain-relevant supporting evidence), which may inform a therapeutic choice~\citep{vincent2013circulatory}.  A model that produces a prediction (e.g., high risk of death) accompanied by clinically meaningful evidence for this prediction  (e.g., poor cardiac health) closely mirrors a clinician's mental model when treating patients, helping to build trust.

In this paper, we present a method to construct predictive models that are accurate \textit{and} provide domain-relevant supporting evidence for predictions (Figure~\ref{fig:model-overview}). The method is designed to provide users that have a deep understanding of the application domain with an appropriate reason to trust or distrust a prediction made by the model. 
Our contributions are:
\begin{enumerate}[nosep,leftmargin=*]
    \item We design a probabilistic model that relates observed data and prediction targets to domain-relevant latent variables. We embed a rich forward model driven by domain knowledge to ensure that the latent variables correspond to domain-relevant concepts. By construction, \textit{maximum a posteriori} (MAP) inference in this probabilistic model yields both accurate predictions and domain-relevant supporting evidence. 
    \item We demonstrate a two-step learning process that approximates MAP inference: (i) estimating parameters of the probabilistic model using variational learning; (ii) approximating MAP estimation of latent variables in the model, yielding a prediction and supporting evidence, using a neural network trained with an objective derived from the probabilistic model. Importantly, we do not need labels for supporting concepts at training time.
    \item We demonstrate our method on a real-world clinical dataset for the task of predicting mortality risk for patients with cardiovascular disease using multimodal electrocardiogram and tabular data. We show that our method produces accurate risk predictions jointly with meaningful supporting evidence for predictions. The supporting evidence captures information that is often only obtained from invasive procedures, and could provide important therapeutic insight for clinicians.
\end{enumerate}

\section{Related Work}

Work on improving trust has often focused on ML model interpretability, including \textit{post hoc} interpretation using input feature attribution~\citep{lundberg2017unified,simonyan2013deep,sundararajan2017axiomatic, zeiler2014visualizing}; \textit{post hoc} concept-based attribution~\citep{ghorbani2019towards,kim2018interpretability}; and constructing explainable models through regularisation~\citep{melis2018towards,plumb2019regularizing} or prototype/input extraction~\citep{al2017contextual,li2018deep,lei2016rationalizing}. Our work differs in spirit from these, since we: (1) do not focus on explaining how the predictive model works, and instead on providing explicit supporting evidence that is useful to domain experts with limited ML understanding; (2) embed a mechanism to provide supporting evidence for predictions directly into the model, rather than relying on \textit{post hoc} analysis; and (3) use prior domain knowledge to inform the higher-level abstractions for supporting evidence, rather than learning abstractions that may not resemble inherently meaningful concepts. 

One closely related work is on Self-Explaining Neural Networks (SENN)~\citep{melis2018towards}, which provide explanations for predictions by forming predictions as a product of input-dependent concepts and weighting terms for these concepts. The concepts for explanations are learned from data (not constrained by domain understanding) and are interpreted by considering input examples that most characterise them. Unlike the supporting evidence we consider, learned concepts in SENN need not resemble meaningful abstractions that a domain expert would find useful in decision making. Finally, SENN uses input examples as prototypes to characterise learned concepts. With complex, multimodal data, as in our clinical experiment, such examples can be challenging to visualise and understand.

Other recent work on using domain-relevant concepts to support predictions is close in spirit to ours~\citep{hind2019ted,koh2020concept,al2017contextual}, but these methods require labelled concepts at training time. In contrast, our work relates the supporting concepts to the predictions through a probabilistic model, such that inference directly yields coupled predictions and supporting evidence for those predictions.  We detail further differences from related work (such as \citet{melis2018towards,koh2020concept}) in the appendix.

\section{Learning to Predict with Supporting Evidence (LPS)}
\label{sec:model}

In this section, we present our method, \textit{Learning to Predict with Supporting Evidence (LPS) }, to construct models that produce both predictions and clinically-relevant supporting evidence.
We define a probabilistic generative model that relates observed data, domain-relevant concepts, and predictive targets, specifying how to ground the concepts using a forward model. \textit{Maximum a posteriori} (MAP) inference in this model jointly yields predictions and domain-relevant supporting evidence. Since such inference is computationally challenging, we present a two-step learning process to approximate MAP inference.

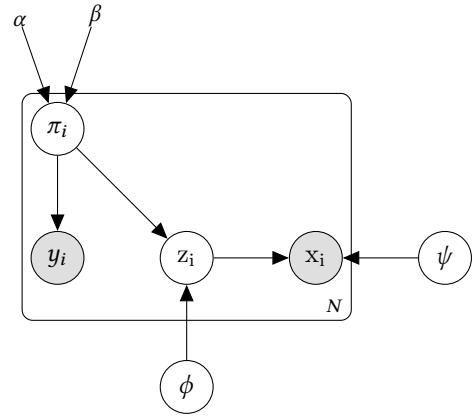
\begin{figure}[t]
\centering
\begin{tikzpicture}

  \node[obs] (y) {$y_i$};
  \node[latent, above=of y] (pi) {$\pi_i$};
  \node[const, above=of pi, xshift=-0.5cm](alpha){$\alpha$};
  \node[const, above=of pi, xshift=0.5cm](beta){$\beta$};
  \node[latent, right=1cm of y]  (z) {$\mathbf{z_i}$};
  \node[obs, right=1cm of z]            (x) {$\mathbf{x_i}$};
  \node[latent, below=1cm of z](phi) {$\phi$};
  \node[latent, right=1cm of x](psi) {$\psi$};

  \edge {pi} {y};
  \edge {pi} {z}; %
  \edge {z} {x};
  \edge {alpha} {pi};
  \edge {beta} {pi};
  \edge {phi}{z};
  \edge {psi}{x};

  \plate {piyx} {(pi)(z)(x)(y)} {$N$} ;

\end{tikzpicture}
\caption{\textit{Probabilistic model for LPS}. The model relates the class label $y_i$ to (latent) class probability vector $\pi_i$, (latent) supporting evidence factors $z_i$, observed features for each data point $x_i$, and distributional parameters $\phi$ and $\psi$. The circled quantities are random variables, with shading used for quantities observed at training time. At inference time, we only observe the features $x_i$. The plate signifies variables that are data point-specific, for $N$ data points, and the quantities outside the plate are common across data points.} 
\label{fig:pgm}
\end{figure}

\subsection{Probabilistic Model}
We describe the model for a general binary classification problem, where the task is to predict class label $y_i$ and positive class probability $\pi_i$, given observed features $x_i$. The model is summarised in Figure~\ref{fig:pgm}. 

Formally, for $i = 1, \ldots, N$, the class label $y_i \in \{0,1\}$ is distributed as \mbox{$y_i\sim \textnormal{Bernoulli}(\pi_i)$} where the positive class probability $\pi_i \in~[0,1]$ is distributed as the flexible prior \mbox{$\pi_i \sim \textnormal{Beta}(\alpha, \beta)$}. Fixed parameters $\alpha$ and  $\beta$ capture the overall balance of the two classes. In clinical risk prediction, $\pi_i$ could represent an individual patient's 60-day risk of death.

We let $z_i \in \mathbb{R}^{m}, z_i \sim p(z_i|\pi_i, \phi)$ be a latent vector encoding interpretable, domain-specific concepts, where $m$ is the number of concepts, and global parameters $\phi$. The variable $z_i$ represents some clinically relevant concept that does not appear directly in the feature space, e.g., estimated glomerular filtration rate (eGFR).
The observed features $x_i \in \mathbb{R}^d, $ are distributed as $x_i \sim~p(x_i | z_i, \psi)$, with global parameters $\psi$. In risk prediction, $x_i$ could be a patient's creatinine, and low creatinine clearance is associated with low eGFR.

\paragraph{\textbf{Domain-Relevant Supporting Evidence:}} We enforce that the variable $z$ encodes domain-relevant concepts by using a well-defined forward model, governed by domain knowledge, that characterises how some subset of the observations $x$ are generated from $z$. For example, the forward model could be a physiological model that relates observable values to latent cardiac function~\citep{catanho2012model}, atlas-based image deformation models~\citep{dalca2019unsupervised}, physics models~\citep{hu2019difftaichi}, or many others. 
We assume a prior $p(\phi)$ determined by domain understanding. 

This modelling directly grounds $z$ to represent domain-relevant concepts by relating $z$ to $x$ via the forward model.  Consistency of $z$ and $\pi$ is enforced through the probabilistic model. 
We emphasise that the forward model need only characterise a sufficient subset of $x$ to constrain $z$; that is, we do not need to define a model to relate $z$ to all of $x$. This is advantageous and broadens LPS's applicability because $x$ may be very high-dimensional and thus it may be challenging to define a forward model governing $x$ in its entirety. 

In this work, we consider only smooth, continuous distributions for the domain knowledge, and a differentiable forward model. In addition to being widely applicable, this enables tractable learning using gradient-based methods.

\subsection{Learning}
\label{sec:learning}
For a given observation $x$, we use \textit{maximum a posteriori} (MAP) estimates of the class label $y$, its associated probability $\pi$, and the supporting concepts $z$:
\begin{align}
    \pi^*, z^*, y^* &= \arg \max_{\pi, z, y} p(\pi, z, y|x) \\
    &= \arg \max_{\pi, z, y} \int p(\pi, z, y|x,\phi,\psi)p(\phi)p(\psi) d\phi d\psi,
\end{align}
Since this is intractable for complex parameter specifications, we approximate it. We use a training set $\mathcal{D} =\nolinebreak \{(x_1, y_1), \ldots, (x_N, y_N)\}$ to learn informative point estimates for the parameters:
\begin{align}
        \phi^*, \psi^* = \arg \max_{\phi, \psi} \log p(\psi, \phi|\mathcal{D}),
\end{align}
and approximate MAP estimates as:
\begin{align}
        \pi^*, z^*, y^* = \arg \max_{\pi, z, y} \log p(\pi, z, y|x, \phi^*, \psi^*).
\end{align}

Since the true posterior \mbox{$p(z,\pi|x, y, \psi, \phi)$} is intractable, we cannot directly use the Expectation Maximization (EM) algorithm to estimate model parameters, because EM requires calculating this posterior exactly. We therefore resort to variational EM~\citep{bernardo2003variational,neal1998view} for parameter estimation: we define a variational approximation 
$q(z,\pi; \theta_q)$ to the true posterior with parameters  $\theta_q$, and then construct a lower bound on the log evidence of data and model parameters as follows. 

First, we express the joint log likelihood of data and model parameters as: 
\begin{align}
\begin{split}
    \log p(\phi, \psi, \mathcal{D}) &= \log p(\phi, \psi, x_1^N, y_1^N) \\
    &= \log p(\phi) + \log p(\psi) +  \\
    &\quad \sum_{i=1}^N\log p(x_i, y_i| \phi, \psi)
\end{split}
\end{align}
Then, we lower bound the data likelihood terms $\log p(x_i, y_i| \phi, \psi)$ with Jensen's inequality (full derivation in appendix), and obtain the following bound:
\begin{align}
    \begin{split}
    \log p(\psi, \phi, \mathcal{D}) &\ge \log p(\psi) + \log p(\phi) +\\
    &\sum_{i=1}^N \Big( \mathbb{E}_{q} \big[ \log p(\pi) + \log p(y_i|\pi) \ + \\
    &\qquad \log p(z|\pi, \phi) + \log p(x_i|z,\psi) \big]  + H(q) \Big), \label{eqn:elbo-vem}
    \end{split}
\end{align}
where $H(q)$ is the entropy of $q$.  We obtain MAP estimates by jointly maximising this lower bound w.r.t. $\phi, \psi,$ and $\theta_q$ using stochastic gradient approximations. This results in MAP estimates $\phi^*, \psi^*$, and an approximate posterior $q(\cdot; \theta_q^*)$.

\subsection{MAP Inference}
For a new data point $x$, we seek the MAP estimates of $\pi, z$, and  $y$:
\begin{align}
    \begin{split}
    \pi^*, z^*, y^* &= \arg \max_{\pi, z, y} \log p(\pi, z, y|x, \phi^*, \psi^*) \\
    &= \arg \max_{\pi, z, y} \log p(\pi) + \log p(y|\pi) + \\
    &\log p(z|\pi,\phi^*) + \log p(x|z, \psi^*).
    \end{split}
    \label{eqn:map}
\end{align}
Depending on the form of each distribution, this maximization might not be solvable in closed form. Computing the MAP estimates on a per-subject basis using EM or gradient ascent is often computationally inefficient with complex $p(x|z, \psi)$. Instead, we use a network $n(x;\theta_n) \rightarrow \nolinebreak (\hat{\pi}, \hat{z})$ with parameters $\theta_n$ to efficiently approximate the MAP estimates such that:
\begin{align}
    \hat{\pi} &\approx \pi^*, \hat{z}\approx z^*, \label{eqn:zpimap} \\
    \hat{y} &= \mathds{1}\left[\hat{\pi} \ge \eta \right] \approx y^*, \label{eqn:ymap}
\end{align}
where $\mathds{1}[\cdot]$ is an indicator function and $\eta$ is a threshold that depends on our desired tradeoff between recall and precision. Equation~\ref{eqn:ymap} follows from the fact that $p(y|\pi)$ is Bernoulli distributed. 

We learn the parameters $\theta_n$ by maximising the MAP objective on the set of training data $\mathcal{D} =\nolinebreak \{(x_1, y_1), \ldots, (x_N, y_N)\}$. Given a network $n(x_i;\theta_n) \rightarrow (\hat{\pi}_i, \hat{z}_i)$, we maximise:
\begin{align}
\begin{split}
    \mathcal{L}_{\textnormal{MAP}}(\theta_n) = \sum_{i=1}^N \Big( &\log p(\hat{\pi}_i) + \log p(y_i|\hat{\pi}_i) + 
    \\ &\log p(\hat{z}_i|\hat{\pi}_i, \phi^*) + \log p(x_i|\hat{z}_i, \psi^*) \Big),
\label{eqn:loss-sl}
\end{split}
\end{align}
which encourages simultaneous accurate risk prediction $\hat{\pi}$ and domain-relevant supporting evidence $\hat{z}$. 
Consistency between the supporting evidence $\hat{z}$ and prediction $\hat{\pi}$ is enforced by the MAP objective from the probabilistic model used to train the inference model (via the term $p(\hat{z}|\hat{\pi})$).

To recover $\pi^{*}, z^{*}, y^{*}$ given a test example $x$, we directly use $n(x;\theta_n^*)$ to obtain the MAP estimates using \eqref{eqn:zpimap}, and \eqref{eqn:ymap}. 

\subsection{Discussion on Modelling}

\paragraph{Representation of domain knowledge:} In some clinical settings, domain knowledge is typically represented using hard constraints or thresholds. For example, one definition of sepsis uses hard thresholds of a severity score~\cite{singer2016third}. However, it is often the case that a continuous distribution is often a more realistic representation of the relationship between relevant concepts and the prediction task. For example, the relationship between severity scores and risk of mortality is well represented as a smooth, continuous function, since higher severity scores correlate with higher risk in a smooth fashion. 

\paragraph{Generality of the model:} This class of models is appropriate for a range of binary classification problems because the Beta distribution is a flexible distribution for $\pi$ and the distributions $p(z|\pi, \phi)$ and~$p(x|z, \psi)$ can be specified as desired. 
Using a Dirichlet prior for $\pi$ and a multinomial for $y$ is a natural extension of this model to the multiclass scenario. In this paper, we focus on binary classification for clinical risk stratification.

\paragraph{Supporting Evidence vs. Explanations:} Since there is no guarantee that there is a causal link between $\hat{z}$ and the prediction $\hat{\pi}$, we refer to $\hat z$ as providing \textit{supporting evidence} for the prediction rather than \textit{explaining} the prediction.

\begin{figure*}[h]
\centering
\includegraphics[height=0.2\textwidth]{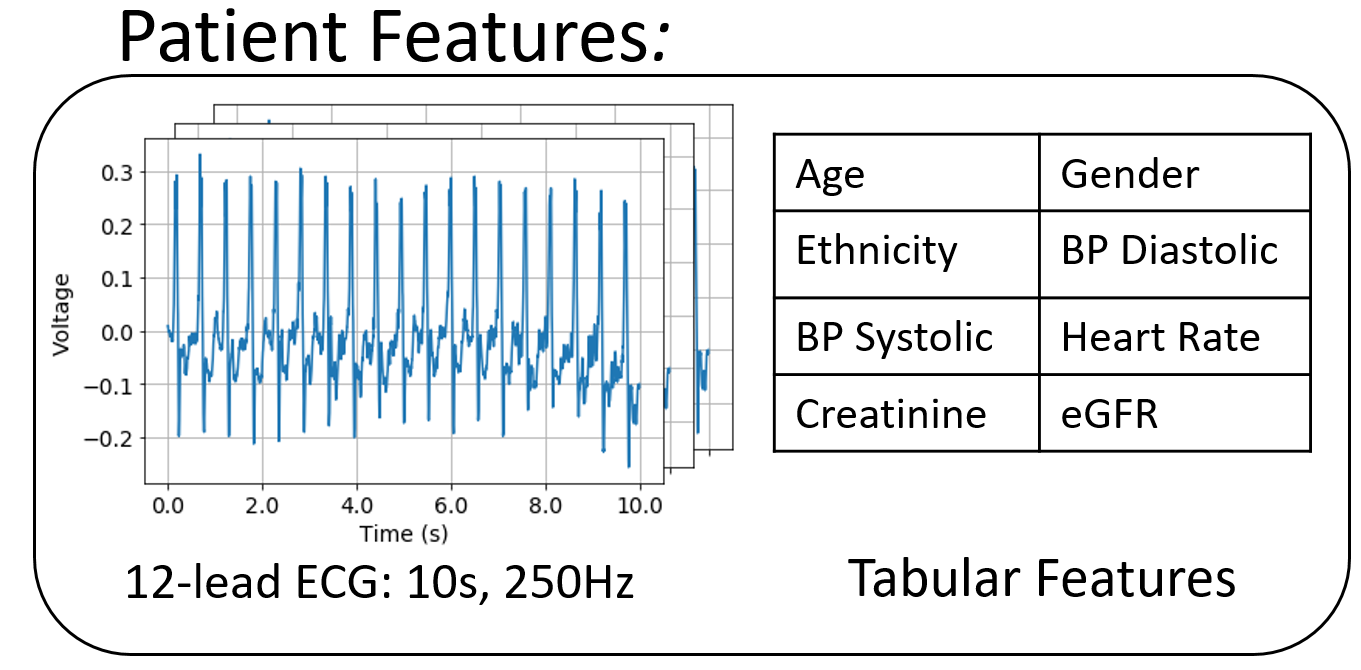}
\caption{Multimodal data for each patient in our clinical dataset: a 12-lead ECG and 8 tabular features.}
\label{fig:apollo-data}
\end{figure*}

\section{Risk Prediction in Real-World Medicine}
\label{sec:med-domain}
We instantiate and demonstrate LPS on the task of predicting a cardiovascular patient's risk of mortality within the 6 months following a cardiac event.  We show that LPS produces accurate risk predictions and informative supporting evidence. Code implementing LPS is available at \texttt{https://github.com/aniruddhraghu/lps}.

Cardiovascular disease affects a large number of people worldwide, and is a major cause of mortality~\citep{benjamin2019heart}. Predicting mortality risk for patients with cardiovascular disease is an important task that has received significant prior study in the medical literature~\citep{granger2003predictors, myers2019,raghunath2020prediction, antman2000timi,morrow2001application,roe2011predicting,mcnamara2016predicting}.  In this section, we focus on predicting patient risk and providing supporting evidence for predictions from electrocardiogram (ECG) and tabular data, which are observed for many cardiology patients in hospital settings.

\subsection{Data}
The dataset we use has 3728 patients from the Massachusetts General Hospital, is de-identified, and was obtained with IRB approval. Each patient has undergone a cardiac catheterisation, which we treat as the index event. Each patient in the dataset has the following measurements and demographic features recorded: 12-lead electrocardiogram (ECG), heart rate (HR), systolic blood pressure, diastolic blood pressure, ethnicity, age, and gender. These values are recorded within 3 days preceding catheterisation event. For each feature, the latest value before the cardiac catheterisation event is used if multiple are available. Additionally, approximately 80\% of the patients have creatinine values recorded within this 3 day period. For those that do not, we use a standard value for a healthy patient (1 mg/dL). From creatinine and the other features, we calculate the estimated glomerular filtration rate (eGFR) for each patient using a standard formula~\citep{levey2009new}. The data are summarised in Figure \ref{fig:apollo-data}.
Each patient has outcome information: survival ($y=0$) or mortality ($y=1$) within 6 months of catheterisation (positive proportion 9.7\%).

\subsection{Model Instantiation}
We let $x$ be the ECG and the tabular features, summarised in Figure \ref{fig:apollo-data}, and $y$ the mortality outcome. The risk $\pi$ is the probability of $y$. We let $z \in \mathbb{R}^5$ be a vector of clinically important quantities: systemic vascular resistance ($R$), arterial compliance ($C$), systole time ($T_s$), diastole time ($T_d$), and cardiac output ($CO$). These quantities are \textbf{not} observed for many patients; for some, such as $R$, $C$, and $CO$, this is because accurate direct measurement is  invasive~\citep{bajorat2006comparison}, and estimating them in (non-invasive) physical exams is challenging~\citep{hiemstra2019diagnostic}. When they are available, they play an important role in risk assessment. For example, when known, the values of $CO$ and $R$ are used to guide the choice of therapeutic interventions for patients with heart failure \citep{yancy20132013}.

\subsection{Probabilistic Model Specification}
We now detail how each distribution in the probabilistic model is specified.
We first define $p(\pi) = \textnormal{Beta}(1.0,1.0)$ and $p(y|\pi) = \textnormal{Bernoulli}(\pi)$. We impose a flat prior on $\psi$:  $p(\psi) \propto 1$.
\paragraph{Specification for $p(z|\pi, \phi)$:} The interpretable latent variable $z$ is 5-dimensional, with components $R, C, T_s, T_d,$ and $CO$. We denote the components of $z$ as $z_m,\  m=1, \ldots, 5$. We factor the distribution $p(z|\pi, \phi)$ as follows:
\begin{align*}
    p(z|\pi, \phi) &= p(R|\pi, \phi_R) p(C|\pi, \phi_C) p(T_s|\pi, \phi_{T_s}) \\
    &p(T_d|\pi, \phi_{T_d}) p(CO|\pi, \phi_{CO}).
\end{align*}
For each $z_m$, we let $p(z_m|\pi, \phi_{z_m})$ have the following mixture form:
\begin{align*}
    p(z_m|\pi, \phi_{z_m})& = \pi \mathcal{LN}(z_m; \mu_{z_m, 1}, \sigma^2_{z_m, 1}) + \\ &(1 - \pi) \mathcal{LN}(z_m; \mu_{z_m, 0}, \sigma^2_{z_m, 0}),
\end{align*}
where $\mathcal{LN}$ is a log normal distribution, such that if $X \sim \mathcal{LN}(\mu, \sigma^2)$, then $\log X \sim \mathcal{N}(\mu, \sigma^2)$.
This mixture form for each component implies the following definition of the parameters: 
$$\phi_{z_m} = \{\mu_{z_m, 1}, \sigma^2_{z_m, 1}, \mu_{z_m, 0}, \sigma^2_{z_m, 0}\}.$$

This instantiation for the distribution has an intuitive interpretation of generating each latent variable $z_m$ for each patient as a mixture of two components, weighted by the patient's risk of death. The two components then correspond to the distribution of the latent feature in the high and low risk cases. 

\paragraph{Specification for $p(\phi)$:}  We define \mbox{$\phi_\mu = \{\mu_{z_m, 0}, \mu_{z_m, 1} \}_{m=1}^5$}, and define \mbox{$\phi_{\sigma^2} = \{\sigma^2_{z_m, 0}, \sigma^2_{z_m, 1} \}_{m=1}^5$}. For each of these means, we impose an independent, normally distributed prior, i.e.: $p(\mu_{z_m, i})  = \mathcal{N}(\tilde{\mu}_{z_m, i}, 0.01^2)$. 
For each of the standard deviations, we specify a delta function prior, i.e.: $p(\sigma^2_{z_m, i})  = \delta( \sigma^2_{z_m, i} -  \tilde{\sigma}^2_{z_m, i})$, which fixes the values of these parameters to be  $\tilde{\sigma}^2_{z_m, i}$.

We now specify how each of these parameters are determined for each component of $z$:
\begin{itemize}[leftmargin=*]
    \item $R$: Take the roughly 20\% of patients in the training dataset that have recorded values for $R$. Consider the subset of these patients that died. For these patients, fit a lognormal distribution to the resulting $R$ values. The parameters of this fitted distribution become $\tilde{\mu}_{R, 1}$ and $\tilde{\sigma}^2_{R, 1}$. Repeat this process for the subset of patients that lived; the resulting parameters are $\tilde{\mu}_{R, 0}$ and $\tilde{\sigma}^2_{R, 0}$.
    \item $C$: No patients have measured values for $C$ (vascular compliance is never directly measured). Approximately estimate the time constant $\tau = RC$ for each patient in the training dataset for which we have $R$ recorded. This is done by using the diastole relation from the two element Windkessel model \citep{catanho2012model} to relate the systolic and diastolic pressures to the time constant. Divide this resulting time constant by $R$ to get approximate values for $C$. Then, determine parameters by following the same process as with determining distributions for $R$.
    \item $T_s$: No patients have measured values for $T_s$. Approximately estimate these by using the fact that $T_s$ is approximately $1/3$ of the duration of a heart beat (reciprocal of heart rate, recorded for all patients). Having estimated these on the training dataset, follow the same process as with $R$ to determine parameters.
    \item $T_d$: No patients have measured values for $T_d$. Approximately estimate these by using the fact that $T_d$ is approximately $2/3$ of the duration of a heart beat (reciprocal of heart rate, recorded for all patients). Having estimated these on the training dataset, follow the same process as with $R$ to determine parameters.
    \item $CO$: follow the same process as with determining $R$, for the $\sim$ 80\% of patients in the training dataset that have recorded values.
\end{itemize}

\paragraph{Specification for $p(x|z,\psi)$:} We partition $x$ into two components: (1) $x^{(g)}$, which is generated from $z$ based on a known forward model $g(z)$; and (2) $x^{(f)}$, which is formed from $z$ based on a forward model $f(z,\psi)$ with parameters $\psi$ that cannot be specified with current domain knowledge, and is thus learned from data.

We let $x^{(g)}$ represent the vital signs (heart rate and blood pressures) and $x^{(f)}$ capture the remaining features. 
We define: $p(x^{(g)}|z, \psi) = \mathcal{N}(g(z), 0.1^2)$, where the known forward model $g(\cdot)$ has two components: (1) the two element Windkessel model to model the blood pressures using a differential equation \citep{catanho2012model, sagawa1990translation,westerhof2009arterial}; and  (2) the definition of the heart rate based on systolic and diastolic times. Concretely, $g(z) \rightarrow (\widehat{BP_{\textnormal{sys}}}, \widehat{BP_{\textnormal{dias}}}, \widehat{HR})$, corresponding to the estimated means for the end systolic pressure, end diastolic pressure, and heart rate respectively. These are produced as follows:

\begin{itemize}[leftmargin=*]
    \item \textbf{Two element Windkessel model:} this is a differential equation model that models the evolution of the blood pressure waveform as a function of the latent variables $R, C, T_s, T_d,$ and $CO$. This model separately characterises the systole phase, where blood is ejected out of the heart through the aorta as a result of ventricular contraction, and the diastole phase, where blood flows into the ventricles at the start of the next cardiac cycle. We adopt the formulation from \citet{catanho2012model}, and the blood pressure in each phase is modelled as follows:
\begin{equation*}
    C \frac{d P(t)}{dt} + \frac{P(t)}{R} = I(t),
\end{equation*}
where
    \begin{equation*}
        I(t) = \begin{cases}
        I_0 \sin\left(\frac{\pi t}{T_s} \right) &\text{during systole}\\
        0 &\text{during diastole.}
        \end{cases}
    \end{equation*}
    $I(t)$ represents the input blood flow, with a half-sinusoid model used for systole, and zero-input for diastole. The constant $I_0$ is set based on the fact that over the systole phase, the entire blood flow must be equal to the stroke volume, $SV$. By definition, stroke volume can be expressed in terms of cardiac output and heart rate $SV = \frac{CO}{HR} = \frac{(T_s + T_d) \times CO}{60}$, and we obtain the following expression for $I_0$:
    \begin{equation*}
       \int_{t=0}^{T_s} I_0 \sin\left(\frac{\pi t}{T_s} \right) dt = SV = \frac{(T_s + T_d) CO}{60}
        \implies I_0 = \frac{\pi \ CO \  (T_s+T_d)}{120 T_s}.
    \end{equation*}
    
    Note that $\pi$ in these equations refers to the mathematical constant, and not the patient's risk of mortality.
    
    With these governing differential equations, we use the piecewise solutions for the blood pressures in each phase from \citet{catanho2012model}. This solution specifies the blood pressure as a function of time. To obtain the desired quantities, $\widehat{BP_{\textnormal{sys}}}$ and $\widehat{BP_{\textnormal{dias}}}$, we step forward the solution for 4 cardiac cycles (each cycle is of duration $T_s + T_d$) so that it reaches steady state. We then evaluate the solution at intervals of $T_s$ and $T_d$ for a further 6 cycles, and average the end systolic and diastolic pressures from these 6 cycles to produce $\widehat{BP_{\textnormal{sys}}}$ and $\widehat{BP_{\textnormal{dias}}}$. Note that the computation of these quantities is differentiable.

    \item \textbf{Definition of heart rate:} the heart rate in beats per minute can be defined as $\frac{60}{T_s + T_d}$. Evaluating this yields $\widehat{HR}$.
\end{itemize}

These two models together fully specify $p(x^{(g)}|z, \psi)$.  We then let $p(x^{(f)}|z, \psi) = \mathcal{N}(f(z,\psi), \sigma_f^2)$, where $f(\cdot)$ is a neural network with parameters $\psi$, which are learned as part of the LPS framework. For the ECG, we let $\sigma_f = 5$, and for the remaining features, we let $\sigma_f=0.5$. More details on network architecture are provided in Section \ref{sec:nn-imp} and in the appendix.

\begin{table*}[t]
\centering
\begin{tabular}{@{}ccccc@{}}
\toprule
Method   &    Thresholded $CO$ F1 Score  &   $R^2(HR,\widehat{HR})$ &   $R^2(BP, \widehat{BP})$ & AUC \\ \midrule
Baseline &    N/A                         &   N/A                     &     N/A                  &  $0.75 \pm 0.02$   \\
SENN     &    N/A                         &   N/A                     &     N/A                  &  $0.73 \pm 0.02$   \\
LPS      &   $\mathbf{0.78 \pm 0.01 }$    &   $\mathbf{0.83 \pm 0.01}$ &$\mathbf{0.90 \pm 0.03}$ &  $0.74 \pm 0.02$  \\
LPS-$q$  &   $0.75 \pm 0.04$              &   $0.78 \pm 0.04$         &     $0.71 \pm 0.02$      &  $0.70 \pm 0.01$  \\ \bottomrule
\end{tabular}
\caption{\textit{LPS predictions and supporting evidence are accurate.}  Analysing the supporting evidence factor $CO$ shows that LPS effectively estimates when this value is below/above a meaningful clinical threshold. Using the forward model to reconstruct the heart rate $\widehat{HR}$ and blood pressure $\widehat{BP}$ from the supporting evidence, the reconstructions capture most of the variance of the true values, suggesting that LPS has captured useful information in this evidence space. In addition, LPS predictions are accurate (comparable AUC to the baseline and SENN \citep{melis2018towards}).  Bolded values are significant at $p < 0.05$.}
\label{tab:apollo-res}
\end{table*}

\subsection{Learning and Inference}
As described in Section \ref{sec:model}, LPS has two stages: firstly, we obtain approximate MAP estimates $\hat{\phi}^*$ and $\hat{\psi}^*$ for model parameters (learning phase); secondly, we train another predictive model to efficiently output MAP estimates for latent model variables (inference phase), yielding a risk prediction and domain-relevant supporting evidence.

We use variational EM to derive approximate MAP estimates $\hat{\phi}^*$ and $\hat{\psi}^*$. We use a deep neural network to model the variational posterior $q(z,\pi|x;\theta_q)$. Again denoting the components of $z$ as $z_m,\  m=1, \ldots, 5$, we use a mean field approximation and factor the variational posterior as $\prod_{m=1}^5 q(z_m|x)q(\pi|x)$.  Each $q(z_m|x)$ is defined to be a log normal distribution, and $q(\pi|x)$ to be a Beta distribution.
The variational posterior network takes the ECG and tabular features as input, and produces mean and variance estimates for the posterior of each $z_m$, and Beta concentration parameter estimates for the posterior of $\pi$.  We approximate the expectation in~\eqref{eqn:elbo-vem} by drawing a single sample from the variational posterior using the reparameterisation trick~\citep{kingma2013auto, rezende2014stochastic}, enabling end-to-end gradient-based training. After running variational EM, we recover approximate MAP estimates $\hat{\phi}^*$ and $\hat{\psi}^*$.

For efficient MAP inference of latent variables, we train a MAP neural network $n(x;\theta_n)$ to take in the ECG and tabular features and directly output MAP estimates of $z$ and $\pi$. This is trained using the objective in Equation \ref{eqn:loss-sl}.

Further details on learning and inference are in the appendix.

\subsection{Neural Network Architectures}
\label{sec:nn-imp}
We now outline the architectures for the three neural networks involved in this LPS instantiation: the learned forward model $f$, the variational posterior $q$, and the MAP inference network $n$.
\begin{itemize}[leftmargin=*]
    \item Learned forward model $f(z,\psi)$: this takes as input $z$ and uses fully connected layers for the tabular features, and a 1D convolutional network with upsampling layers for the ECG.
    \item Variational posterior $q(z,\pi|x;\theta_q)$: this takes as input the ECG and tabular features, $x$. The ECG is passed through a 1D convolution residual network, and the tabular features through a two layer fully connected network. These representations are concatenated and passed through additional fully connected layers to produce mean and variance estimates for the posterior of each $z_m$ and Beta concentration parameter estimates for the posterior of $\pi$.
    \item MAP inference network $n(x; \theta_n)$: this is a neural network with the same architecture as the variational posterior network, except that it directly outputs the MAP estimates of $z$ and $\pi$, rather than distributional parameters.
\end{itemize}

Further architectural and training details are in the appendix.

\subsection{Experiment Details}

\paragraph{\textbf{Performance Baselines:}} We train a network with the same architecture as $n(x;\theta_n)$ to predict only the class label without the supporting evidence. We use this to investigate whether simultaneously learning to predict and supporting evidence impacts the quality of the prediction. We use SENN~\citep{melis2018towards} as a second baseline, which both predicts and provides information designed to supplement the prediction. 

\paragraph{\textbf{Ablation:}} As an ablation, we consider a variant of LPS that we call LPS-$q$. Here, instead of training a separate MAP inference network $n(x;\theta_n)$, we take the variational posterior $q(z,\pi|x;\theta_q)$ and use the mode of this posterior to obtain MAP estimates for $z$ and $\pi$. The modes of the respective log normal and Beta distributions have simple analytical forms in terms of the distributional parameters that are output by the variational posterior network, so are easy to compute. This simpler variant does not require training a separate MAP inference network.

\paragraph{\textbf{Evaluation:}} We use the median/half IQR for ten runs, splitting the dataset into ten train/validation/test sets (60\%/20\%/20\%), and Welch's $t$-test for statistical significance.

\begin{figure*}[t]
\centering
\centerline{
  \begin{tabular}{ccc}
    \includegraphics[width=0.33\linewidth]{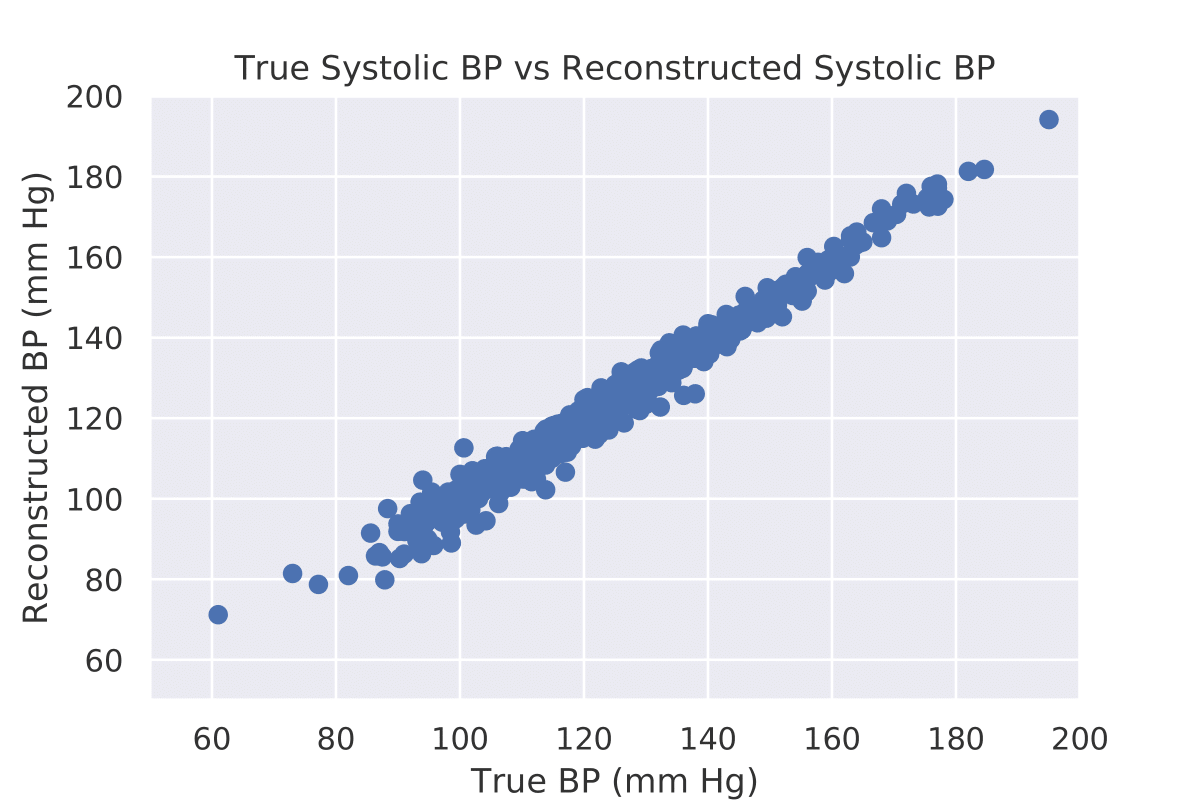}   & 
    \includegraphics[width=0.33\linewidth]{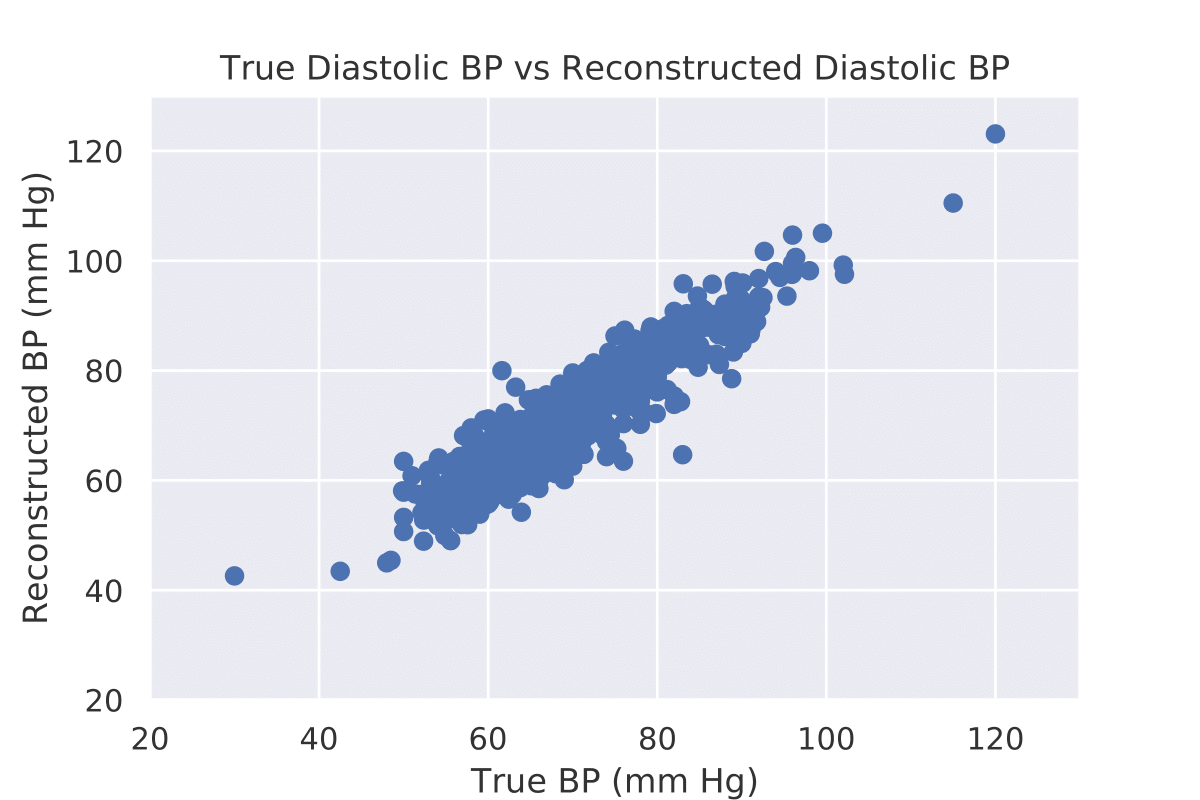} &  
    \includegraphics[width=0.33\linewidth]{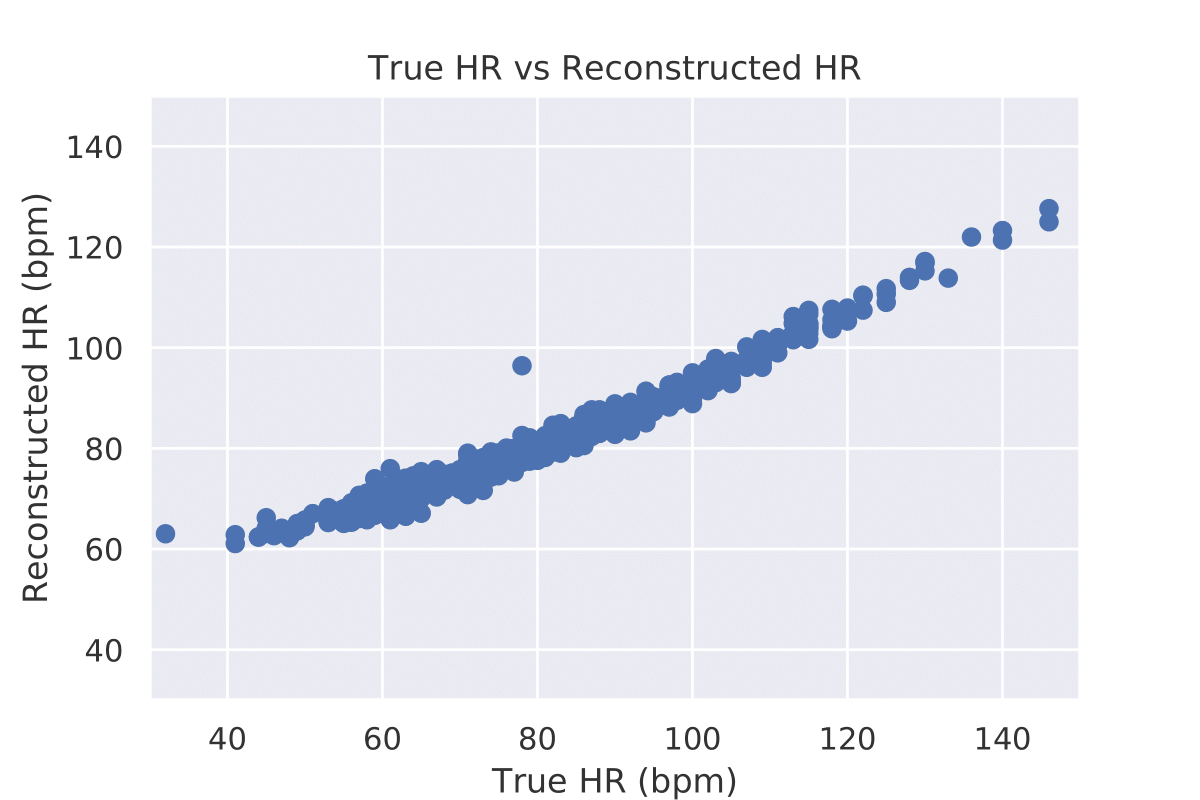} 
  \end{tabular}}
  \caption{ \textit{Reconstructing observed tabular features using the forward model.} We visualise the true blood pressures and heart rate vs. the reconstructed blood pressures and heart rate calculated from the inferred latent concepts using the forward model. The true and reconstructed quantities show good agreement, indicating successful recovery of the latent factors.}
 \label{fig:scatter-expl}
\end{figure*}

\begin{figure*}[t]
\centering
\centerline{
  \begin{tabular}{cc}
    \includegraphics[height=0.2\linewidth]{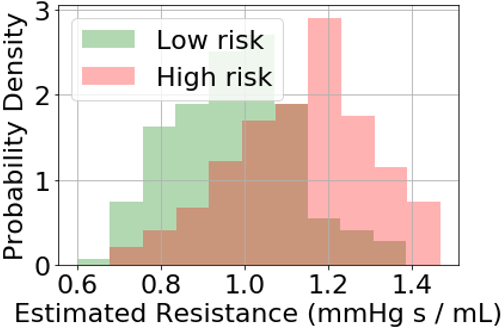}   & 
    \includegraphics[height=0.205\linewidth]{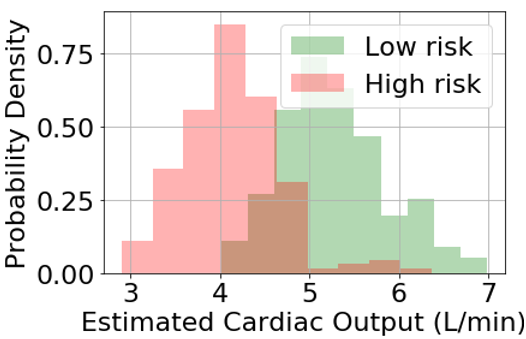}
  \end{tabular}}
  \caption{\textit{Supporting evidence concepts and risk predictions have distributions that are in alignment with domain understanding.} Empirical distributions (histograms) of clinically meaningful factors for patients in the upper and lower quartiles of predicted risk are in accordance with clinical domain knowledge. Such agreement is important for supporting evidence and predictions to be trusted by clinicians \citep{stultz2019advent,tonekaboni2019clinicians}.}
  \label{fig:apollo-concept-risk}
\end{figure*}

\subsection{Results}
\begin{figure*}[ht]
\centering
\centerline{
  \begin{tabular}{c}
    \includegraphics[width=0.9\linewidth]{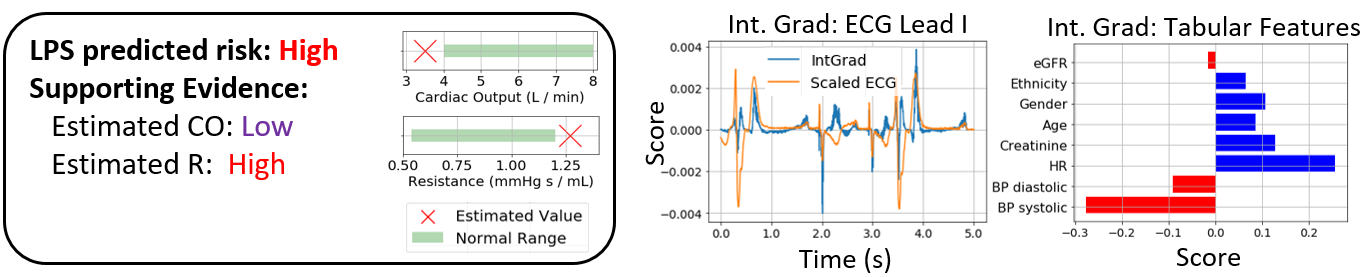} \\ 
    \includegraphics[width=0.9\linewidth]{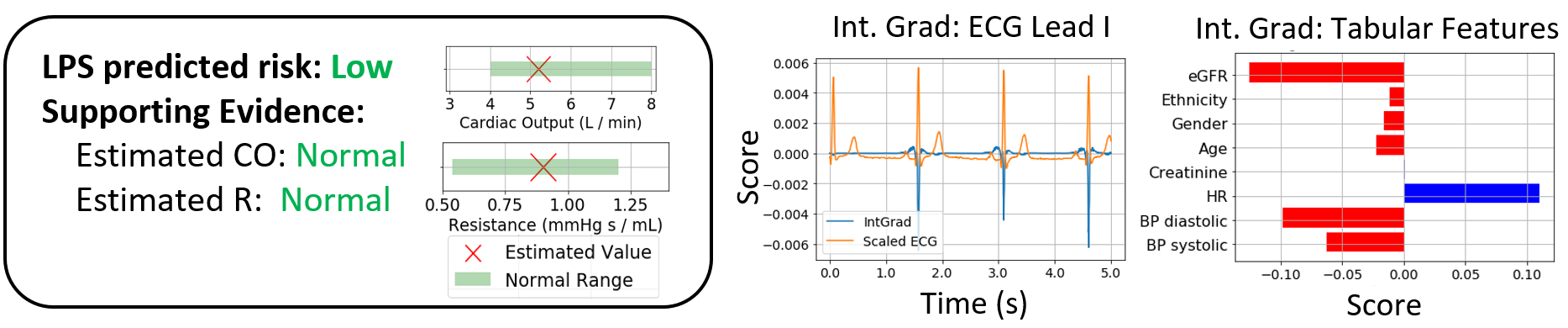}
  \end{tabular}}
  \caption{ \textit{Supporting evidence from LPS is comprehensible and provides actionable insights.} In comparison to attribution methods such as Integrated Gradients, LPS produces supporting evidence that is clinically meaningful and provide insights beyond the input feature space. For a high risk patient (top), LPS produces actionable supporting evidence, namely that that $CO$ and $R$ (which are hard to observe and important in therapeutic decisions) lie outside their normal ranges. For a low risk patient (bottom), LPS recovers supporting evidence factors within their normal ranges. This supporting evidence for a prediction is actionable and less ambiguous than feature attribution methods such as Integrated Gradients, especially when applied to a high-dimensional input such as the ECG.}
  \label{fig:expl-apollo-y1}
\end{figure*}

\begin{figure*}[t]
\centering
  \centerline{\includegraphics[width=0.9\linewidth]{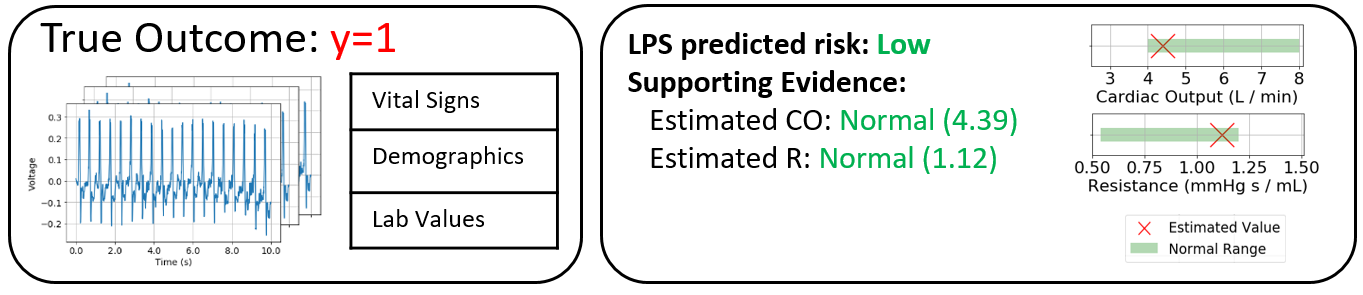}}
  \caption{ \textit{LPS supporting evidence could provide actionable insights on challenging, borderline cases.} The patient shown had an adverse outcome ($y=1$), but was predicted by LPS to be at low risk -- a misclassification. Analysing the weak supporting evidence for this decision adds more insight. 
  }
  \label{fig:apollo-expl-misclf}
\end{figure*}

Table \ref{tab:apollo-res} summarises the results on the clinical dataset. While we focus on analysing the supporting evidence, we also find that LPS, SENN, and the baseline perform comparably in terms of AUC, and that LPS-$q$ performs worse than LPS, justifying the use of the MAP inference network.  This is an expected result, since the variational posterior must model the entire posterior distribution over latents, and not just the modes. Thus, it may sacrifice accuracy in recovering the modes to better represent the distribution as a whole. 

\paragraph{\textbf{Accuracy of supporting evidence.}} We first compare the accuracy of the supporting information for the $\sim 80\%$ of patients that have $CO$ measurements. We compare how the estimated $CO$ compares to the measured $CO$ by splitting $CO$ into two groups using a cutoff of $4$ L/min, which corresponds to the lower limit of normal range for $CO$~\citep{hurst1990heart} -- a clinical standard that informs practice. Computing the resulting F1 score, we observe that LPS estimates do a good job of differentiating between patients who have low and normal cardiac outputs (Table \ref{tab:apollo-res}). Since $CO$ is usually estimated using invasive procedures and is important in clinical decision making, the fact that LPS can (non-invasively) identify when $CO$ is above/below a meaningful threshold is clinically valuable.

Too few patients have measurements of the other latent variables to enable a meaningful direct comparison. In lieu of doing this, we use the known forward model to reconstruct both the heart rate $\widehat{HR}$ (from $T_s$ and $T_d$) and the blood pressures $\widehat{BP}$ from the estimated latent values. We observe high coefficient of determination between the reconstructed quantities and the measured values for HR and BP (Table \ref{tab:apollo-res}). This suggests effective recovery of the latent parameters. 

As further analysis, Figure \ref{fig:scatter-expl} visualises the true blood pressures and heart rate, and the reconstructed estimates from the forward model using the inferred latent concepts from the model. As can be seen, these show good agreement, indicating successful recovery of the latent factors. The median absolute error in recovering quantities was: BP-systolic: $1.62$ mmHg; BP-diastolic: $2.92$ mmHg; HR: $4.75$ bpm.

\paragraph{\textbf{Consistency of predictions and supporting evidence.}}  Figure \ref{fig:apollo-concept-risk} shows histograms for the latent concepts for patients at high and low predicted risk (top and bottom 25\% respectively). On average, patients at the highest risk of adverse outcomes have a lower cardiac output and higher systemic vascular resistance relative to those who do not have adverse outcomes~\citep{thenappan2016critical}.  This is recovered by LPS, achieving consistency of predictions and supporting evidence with clinical domain knowledge. 

\paragraph{\textbf{Comparing supporting evidence.}} Figure \ref{fig:expl-apollo-y1} shows supporting evidence for model decisions from LPS and attributions from Integrated Gradients \citep{sundararajan2017axiomatic}, a commonly used feature attribution method, on the baseline. Normal ranges for the concepts were derived from literature~\citep{klingensmith2008washington, saouti2010arterial,thenappan2016critical}. 
We focus on showing data for $CO$ and $R$, since these are most meaningful to clinicians and there are therapeutic interventions for modifying them~\citep{yancy20132013}. We compare LPS and Integrated Gradients for two patients, one at high risk of death (top), and one at low risk of death (bottom).

LPS produces supporting evidence that is clinically understandable, specifically that the patient shown at the top is at high risk and has elevated $R$ and low $CO$. These statements give insights beyond the feature space alone. In contrast, the baseline of Integrated Gradients reveals certain features in the tabular data and the ECG that contributed to the decision, but the attributions are not as readily actionable as the supporting evidence from LPS. For example, the patient in Figure~\ref{fig:expl-apollo-y1} had a normal heart rate (HR) of 80 bpm (normal range: 60-100 bpm), yet the HR feature contributed significantly to the prediction of high risk. It is unlikely that knowing that the HR was a factor in the model's prediction would lead to any action on the part of a clinician. 

For the low risk patient (bottom), LPS's supporting evidence for a prediction of low risk is accompanied by the inference that the patient has normal $CO$ and $R$. Integrated gradients analysis reveals that most of the tabular features and the QRS complex of the ECG contribute to a low risk prediction, but the HR, recorded as 60 bpm, elevates the patient's risk. For a clinician who is not an ML practitioner, it may be challenging to disentangle these different factors and thus understand/trust the model's predictions using the Integrated Gradients attribution.

\begin{figure*}[t]
\centering
  \centerline{\includegraphics[width=0.9\linewidth]{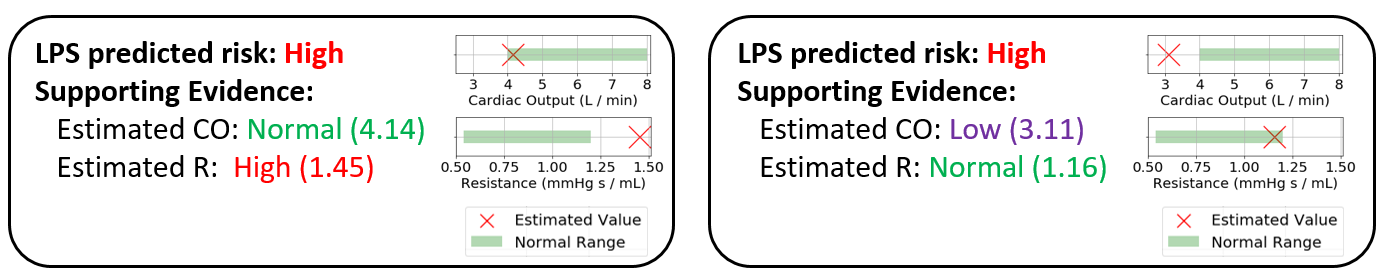}}
  \caption{ \textit{LPS supporting evidence offers \textit{patient-specific insights} to clinicians.} For two patients who were predicted to be at high risk ($\hat{\pi} > 0.8$) LPS supporting evidence captures different factors contributing to risk, potentially informing clinicians of the most suitable \textit{patient-specific} medical interventions.}
  \label{fig:apollo-expl-comp}
\end{figure*}

\paragraph{\textbf{LPS supporting evidence could provide therapeutic insight on challenging examples.}} In a case where LPS makes a misclassification on a hard example, Figure \ref{fig:apollo-expl-misclf}, the weakness of the supporting evidence could be clinically informative. LPS predicted that this patient is at low risk, yet they had an adverse outcome (a misclassification). The supporting evidence is that the patient had normal range values for $R$ and $CO$, which are only just inside the  normal range. This could inform the clinician that although the patient was predicted to be at low risk, the patient might benefit from additional monitoring. Such explanatory insights are challenging to obtain from existing explainability methods.

\paragraph{\textbf{Patient-specific insights from supporting evidence.}} Figure ~\ref{fig:apollo-expl-comp} compares supporting evidence for a pair of patients predicted to be at high risk. The predictions and supporting evidence are clinically meaningful: on the left, the patient has high $R$, and on the right, the patient has low $CO$, both of which are typically indicative of poor cardiovascular health~\citep{hurst1990heart,thenappan2016critical}. LPS supporting evidence is well-differentiated on a per-patient basis and could inform a clinician that the left patient could benefit from medication to reduce $R$, and the right patient could benefit from medication to increase $CO$.


\section{Conclusion}
To assist human experts in decision making, machine learning models should produce both accurate predictions and supporting evidence for these predictions. In healthcare, this consideration is particularly important since clinicians draw significantly on medical principles in their decision making, and therefore can act most effectively on predictions that are accompanied by clinically relevant supporting evidence. 

To tackle this problem, we propose a method, \textit{Learning to Predict with Supporting Evidence (LPS)}, to construct models that provide both predictions and supporting evidence using clinically-relevant concepts. We demonstrate that LPS produces accurate predictions and comprehensible supporting evidence for predictions on a real-world medical dataset. 

LPS relies on domain knowledge to inform (1) the choice of concepts for supporting evidence, and (2) how these concepts are related to the observed data and the prediction.
In medicine, there is a tremendous amount of domain knowledge of this form, obtained from decades of medical practice and modelling. This includes probabilistic models for imaging and disease progression, physiological models for signals, and more~\citep{catanho2012model,dalca2019unsupervised,westerhof2009arterial}.
LPS could be applied to other prediction problems by leveraging such medical domain knowledge, helping to further the trustworthiness and actionability of machine learning models for healthcare.

\section*{Ethics Statement}
This work represents an initial step towards improving trustworthiness in ML models using supporting evidence. More testing of the method would be necessary prior to any real-world deployment. If the supporting evidence produced by the method ends up misleading care givers it could lead to poor decisions. It is important that prior to any deployment, extensive user studies are performed in order to detect such issues and prevent potential negative impact.

\section*{Acknowledgements}
This work was supported in part by funds from Quanta Computer, Inc. The authors thank the members of the Clinical and Applied Machine Learning group and the Computational Cardiovascular Research group at MIT for all their helpful comments and advice. 
\bibliographystyle{ACM-Reference-Format}
\bibliography{example_paper}
\clearpage
\onecolumn
\appendix


\section{Derivation of lower bound for variational EM}
\label{sec:app-vem-derivation}

We lower bound the joint log likelihood of data and model parameters:
\begin{align}
    \log p(\phi, \psi, \mathcal{D}) &= \log p(\phi, \psi, x_1^N, y_1^N) \\
    &= \log p(\phi) + \log p(\psi) + \log p(x_1^N, y_1^N| \phi, \psi).
\end{align}

Consider the data likelihood term alone:

\begin{align}
    \log p(x_1^N, y_1^N| \phi, \psi) &= \sum_{i=1}^N \log p(x_i, y_i|\phi, \psi),
\end{align}

because data points are iid.

Then, considering a single term of this sum:
\begin{align}
&\log p(x_i, y_i|\phi, \psi) \\
&= \log \Big( \int p(\pi, y_i, z, x_i|\phi,\psi) d\pi dz \Big) \\
                            &= \log \Big(\int \frac{q(\pi, z)}{q(\pi, z)} p(\pi, y_i, z, x_i|\phi,\psi) d\pi dz \Big) \\
                            &\ge \int q(\pi, z) \log \Big(  \frac{p(\pi, y_i, z, x_i|\phi,\psi)}{q(\pi, z)}\Big) d\pi dz \label{eqn:jensen} \\
                            &= \mathbb{E}_{\pi, z \sim q(\pi, z)} \Big[ \log p(\pi, y_i, z, x_i|\phi, \psi) \Big] + H(q),
\end{align}
where $q$ is some distribution over the latent variables $\pi, z$, the inequality in (\ref{eqn:jensen}) comes from Jensen's inequality and concavity of $\log$, and $H(q)$ is the entropy of distribution $q$.

Then, consider the term inside the expectation:

\begin{align}
    &p(\pi, y_i, z, x_i|\phi, \psi)  \\
    &=  \frac{p(\pi, y_i, z, x_i,\phi, \psi)}{p(\phi, \psi)} \\
    &= \frac{p(\phi) p(\psi) p(\pi) p(y_i|\pi) p(z|\pi, \phi) p(x_i|z, \psi)}{p(\phi)p(\psi)} \\
    &= p(\pi) p(y_i|\pi) p(z|\pi, \phi) p(x_i|z, \psi).
\end{align}

We then arrive at the final result:
\begin{align}
\begin{split}
    &\log p(\phi, \psi, \mathcal{D}) \ge \\
    &\quad \Bigg( \log p(\phi) + \log p(\psi) + 
    \sum_{i=1}^N \mathbb{E}_{\pi, z \sim q(\pi, z)} \Big[ \log p(\pi) + \log p(y_i|\pi) + 
    \log p(z|\pi, \phi) + \log p(x_i|z, \psi) \Big] + H(q) \Bigg).
\end{split}
\end{align}

\newpage
\section{Additional details on Related Work}
We provide more details comparing LPS to three methods that support predictions with concept-based explanation:

\begin{itemize}
    \item Contextual Explanation Networks \citep{al2017contextual}: Uses both high dimensional input (e.g. an image) and a set of labelled attributes for each example in making a predictive decision, with the high-dimensional input used to generate weights for these attributes in the predictor. In contrast, LPS does not assume that these labelled attributes exist for every example, and instead enforces groundedness of the supporting evidence concepts using a forward model and domain knowledge. This assumption of labelled attributes is why CEN is not used as a baseline method in the evaluation of LPS.
    \item Self-Explaining Neural Networks \citep{melis2018towards}: provides explanations for predictions by learning a neural network model that forms predictions as a product of input-dependent concepts and weighting terms for these concepts. The concepts for explanations are learned from data (not constrained by domain understanding) and are interpreted by considering input examples that most characterise them (they do not necessarily have an inherent interpretation). Unlike concepts in LPS, learned concepts in SENN need not resemble meaningful abstractions that a domain expert would find useful in decision making. Furthermore, since SENN uses input examples as prototypes to characterise learned concepts. With complex, multimodal data, as in our clinical experiment, such examples can be challenging to visualise and understand, unlike LPS concepts, which have a direct interpretation.
    \item Concept Bottleneck Models \citep{koh2020concept}: Outputs a higher level set of concepts along with a prediction, in a similar fashion to LPS. However, CBMs assume labels for each concept at training time, whereas LPS relies on domain knowledge to ground the concepts. Technically, LPS uses a generative modelling approach to incorporate domain knowledge at training time, whereas CBMs do not.
\end{itemize}
 
 \section{Additional Information for Experiments}
We provide further information for experiments.
\subsection{Network Architectures}
\paragraph{Learnable forward model:} The network $f(z,\psi)$ models observed tabular features and the ECG. Generating the entire 12 lead ECG is challenging, so we approximate the respective term in the objective function ($p(x^{(f)}|z, \psi)$) by computing the log probability of the first $500$ samples of the first lead. The network architecture is as follows:
\begin{itemize}[leftmargin=*]
    \item ECG network:
    \begin{itemize}
        \item Input: $z$
        \item FC layer, output size 100, leaky ReLU activation
        \item Upsample (scale factor 1.5), 1D conv (32 channels, kernel size 15, stride 1), Batch norm, leaky ReLU
        \item 1D conv (64 channels, kernel size 15, stride 1), Batch norm, leaky ReLU
        \item Upsample (scale factor 1.5), 1D conv (128 channels, kernel size 15, stride 1), Batch norm, leaky ReLU
        \item 1D conv (128 channels, kernel size 15, stride 1), Batch norm, leaky ReLU
        \item Upsample (scale factor 1.33), 1D conv (64 channels, kernel size 15, stride 1), Batch norm, leaky ReLU
        \item Upsample (scale factor 1.33), 1D conv (32 channels, kernel size 15, stride 1), Batch norm, leaky ReLU
        \item Upsample (scale factor 1.25), 1D conv (1 channel, kernel size 15, stride 1)
    \end{itemize}
    \item Tabular features network:
    \begin{itemize}
        \item Input: $z$
        \item FC layer, output 128,  Batch norm, leaky ReLU
        \item FC layer, output 128,  Batch norm, leaky ReLU
        \item FC layer, output 64,  Batch norm, leaky ReLU
        \item FC layer, output 7,  Batch norm, leaky ReLU.
    \end{itemize}
\end{itemize}

\textbf{Variational posterior:} We use a deep neural network to model the variational posterior, represented as $q(z,\pi|x;\theta_q)$. Using a mean field approximation we factor this posterior as $\prod_{m=1}^5 q(z_m|x)q(\pi|x)$. We set each $q(z_m|x)$ to be a lognormal distribution, and $q(\pi|x)$ be a Beta distribution.

The network architecture is follows:
\begin{itemize}[leftmargin=*]
    \item The ECG is passed through a 1D convolution residual network, based on the ResNet-18 architecture. This network uses kernel of size 15 throughout, with 4 blocks of 32, 64, 128, and 256 channels respectively. Each block downsamples the input by a factor of 2.
    \item The tabular features are passed through a two layer FC network with ReLU activation and 64 and 128 hidden units.
    \item The representation from the ECG is average pooled in the temporal dimension, and is concatenated with the representation from the tabular features.
    \item This is passed through 2 more FC layers, with ReLU activation, and 128 and 64 hidden units respectively. 
    \item A final FC layer produces the mean and variance parameters for the posterior on each $z_m$, and the Beta distribution concentration parameters for the posterior on $\pi$. The variance estimates have exponential function activation. The Beta concentration parameters are clamped softly using a sigmoid function to be in the range $[1, 11]$ for numerical stability.
\end{itemize}

\textbf{MAP Inference Network:} The MAP network $n(x;\theta_n)$ is a neural network with the same architecture as the variational posterior network, except that it directly outputs the MAP estimates of $z$ and $\pi$.


\subsection{Implementation and Training Details}
We learn MAP parameter estimates for the model by maximising a lower bound on the log evidence. We approximate the expectation in this lower bound by drawing a single sample from the variational posterior using the reparameterisation trick~\citep{kingma2013auto, rezende2014stochastic}, allowing end-to-end gradient-based training of the parameters $\phi$, the variational posterior parameters $\theta_q$ and the forward model parameters $\psi$. 

In practice, we use a small number of empirical adjustments: (i) we only maximise the log probability of the first 500 samples of the ECG (to simplify the modelling problem); (ii) for the first 10 epochs of training, we do not use the data likelihood term $p(x|z,\psi)$ in the objective, for learning stability. 

The objective function written out in finite sample form, for a batch size of $K$ points is:

\begin{align*}
    &\mathcal{L}(\phi, \psi, \theta_q) = \\
    &\Bigg( \log p(\phi) + \log p(\psi) + 
    \sum_{i=1}^K \Big[ \log p(\tilde{\pi}) + \log p(y_i|\tilde{\pi}) +
    \log p(\tilde{z}|\tilde{\pi}, \phi) + \log p(x_i|\tilde{z}, \psi) - 
    \log q(\tilde{z};\theta_q) - \log q(\tilde{\pi};\theta_q) \Big] \Bigg),
\end{align*}
with $\tilde{\pi}, \tilde{z} \sim q(z|x_i;\theta_q) q(\pi|x_i;\theta_q)$.

We use Adam \citep{kingma2014adam} for maximising this lower bound, with a learning rate of 1e-4. Training is for 200 epochs with a batch size of 32. This yields approximate MAP estimates $\phi^*, \psi^*$, and variational posterior parameters $\theta_q^*$.

To learn the MAP inference network $n(x;\theta_n) \rightarrow (\hat{\pi}, \hat{z})$, we train with Adam for 200 epochs, with a learning rate of 1e-3 and a batch size of 32. The objective function is as follows, with a batch size of $K$:
\begin{align*}
    \mathcal{L}_{\textnormal{MAP}}(\theta_n) = \sum_{i=1}^K \Big[ \log p(\hat{\pi}) + \log p(y_i|\hat{\pi}) + 
    &\log p(\hat{z}|\hat{\pi}, \phi^*) + \log p(x_i|\hat{z}, \psi^*)\Big].
\end{align*}

The baseline model has the same architecture as the MAP inference network, except it only outputs the class probability and is trained using a standard binary cross entropy loss. This model is trained for 200 epochs with Adam, with a learning rate of 1e-3 and batch size of 32.
SENN is implemented with the same autoencoder architecture as the LPS variational autoencoder, and uses 5 basis concepts (to enable comparison with LPS). We examined different learning rate and sparsity parameters on the validation set.

\textbf{Choice of architectures and hyperparameters:} For the learnable forward model, we based our architecture on standard upsampling architectures used in deconvolutional networks. We investigated shallower and deeper architectures and decided on this architecture based on reconstruction performance. 
The baseline, MAP, and variational posterior networks here using FC layers and 1D CNNs is based on the architecture from \citep{raghunath2020prediction}, with an additional residual network structure for the backbone of the network to extract ECG features. These architectures were not tuned.

We compared learning rates of 1e-3, 5e-4, and 1e-4 for all models, and decided on the final learning rate based on stability of training. We did not tune other model hyperparameters (batch size, weighting of loss terms, etc).

\textbf{Early stopping:} Validation set AUC performance was used to decide when to evaluate on the test set, for all models.

All models were implemented in PyTorch and trained on an NVIDIA Titan Xp GPU.

\end{document}